\documentclass[runningheads]{llncs}

\usepackage[T1]{fontenc}
\usepackage{graphicx}
\usepackage{amsfonts}
\usepackage{amsmath}
\usepackage{booktabs}
\usepackage{multirow}
\usepackage{microtype}
\usepackage{xcolor}
\usepackage{url}
\usepackage{marvosym}

\makeatletter
\@ifundefined{credits}{}{}
\makeatother

\begin{document}

\title{GAFT: Geo-Anchored Fine-Tuning for\\Hazard Identification from Rare Failures}

\titlerunning{Geo-Anchored Fine-Tuning}
\authorrunning{Y. Xu et al.}

\author{Yanran Xu\textsuperscript{\Letter}, Chuanhang Qiu, Yue Wang, Wenbo Wu, Zhaoxing Li}
\institute{University of Southampton, Southampton, United Kingdom \\ \email{\{Y.Xu, C.Qiu, Yue.Wang, Wenbo.Wu, Zhaoxing.Li\}@soton.ac.uk}}

\maketitle

\begin{abstract}
Off-road navigation can fail when physical structures induce irrecoverable states such as high-centering or entrapment, requiring human interventions. Identifying these structures is crucial, yet challenging. Such failure events are rare and costly to collect, resulting in limited training data. Moreover, the collected data associate frames with outcomes, but do not indicate the visual cues responsible for the failure. Learning directly from these data can therefore exploit scenario-specific visual cues, leading to poor generalization. We propose \textbf{Geo-Anchored Fine-Tuning (GAFT)}, a parameter-efficient method that adapts a vision foundation model with a geometry-derived prior. It guides LoRA adaptation by aligning a spatial attention-rollout map with the geometry prior, while preserving pretrained representations. On an intervention-verified forest hazard benchmark, across ten independently trained adaptations, GAFT consistently outperforms frozen DINOv2 and supervised PEFT baselines, improving the repeated leave-one-scenario-out mean $F_2$ from 0.0607 to 0.3757 with statistical significance under paired analysis. Within these independently trained models, the best-performing GAFT model achieves a repeated-LOSO $F_2$ of 0.570. Code and benchmark: \url{https://github.com/Xu-Yanran/geo_anchored_fine_tuning}.

\end{abstract}


\section{Introduction}

For autonomous ground robots operating in unstructured terrain, the presence of a physical structure alone does not determine whether it poses a hazard.  The risk posed by a structure depends on its geometry relative to the robot and its path.  A standing tree trunk may be visually prominent yet remain avoidable. Conversely, a low fallen log can high-center a rover that drives over it, causing loss of wheel--ground contact.  We use \emph{hazards} to denote structures that can drive a ground robot into such irrecoverable states, including high-centering, entrapment, and sensor damage \cite{jinResilientLeggedLocal2023,xu2025bw}.  Reliable field navigation therefore requires reasoning about the physical consequences of local structures, beyond their object identity or visual saliency.

Learning this distinction is difficult because hazard data are scarce and provide only coarse, outcome-level supervision. Hazard events are costly to collect, often requiring human intervention after a failure, resulting in only a small number of labeled examples. Moreover, these data associate frames with failure outcomes but do not indicate the visual evidence responsible for the hazardous interaction. Consequently, models trained directly on such data, whether from scratch or by adapting a pretrained model using only hazard labels, may exploit scenario-specific cues, such as scene layout, texture, illumination, or obstacle appearance, rather than learning the robot–terrain relationship that makes a structure hazardous \cite{geirhos2020shortcut}.  Few-shot and meta-learning methods reduce annotation requirements when suitable meta-training tasks are available~\cite{finn2017model,snell2017prototypical,xue2025fsl}. Such tasks are difficult to construct for rare off-road failures spanning diverse robots, terrains, and obstacle configurations. Class-imbalance objectives address label imbalance~\cite{lin2017focal,qiu2025smote} but do not constrain the evidence learned under outcome-level supervision, leaving shortcut learning unresolved.

Vision foundation models (VFMs) provide a strong starting point because their pretrained features capture broad visual and semantic structure without learning a representation from scarce hazard data alone. Wild visual navigation uses pretrained DINO-ViT features to learn traversability from field experience \cite{mattamala2025wild}, and self-supervised ViTs provide transferable patch representations~\cite{caron2021emerging,oquab2024dinov2}. However, generic visual pretraining provides no preference for features that predict the physical consequence of a particular robot--terrain interaction. A VFM may encode trees, foliage, terrain texture, and scene layout well; when it is adapted only on scarce hazard data, a classifier or parameter-efficient adapter can therefore rely on whichever of these cues best predicts the observed scenarios, including scene-specific shortcuts that do not transfer. Data-scarce hazard identification consequently requires an additional inductive bias that preserves transferable visual features while favouring spatial cues relevant to traversal.

We propose Geo-Anchored Fine-Tuning (GAFT), a two-stage parameter-efficient method that adapts a vision foundation model for hazard identification from rare failures. GAFT uses an RGB–depth traversal dataset to construct a proximity-weighted geometric prior that emphasizes structures protruding above the ground, assigning greater weight to those closer to the robot. During adaptation, attention rollout is aligned with this prior while a representation anchor limits drift from the pretrained representation. The hazard classifier is then trained on scarce RGB frames whose labels are derived from the observed outcomes of robot-terrain interactions.

\noindent\textbf{Contributions:}
\begin{enumerate}
  \item We introduce a two-stage learning framework that uses an RGB–depth traversal dataset to guide vision foundation model adaptation before training a robot-specific hazard classifier from scarce robot–terrain interaction labels.
  \item We propose GAFT, a parameter-efficient adaptation method that aligns spatial attention with a proximity-weighted geometric prior to guide an RGB vision foundation model while preserving pretrained representations.
  \item Empirical evaluation on an intervention-verified benchmark shows that GAFT consistently outperforms frozen DINOv2 and supervised PEFT baselines, with statistical significance under paired analysis.
\end{enumerate}

\section{Related Work}

\subsection{Navigation in Unstructured Environments}
Navigation in unstructured environments has often been approached by estimating traversability from the geometry of the surrounding terrain.  Geometry-based methods capture slope, clearance, and other structural properties relevant to navigation~\cite{chavez2018learning,frey2022locomotion}.  However, geometry is only a partial description of traversability: tall grass may be traversable despite its geometric extent, while visually inconspicuous structures near the ground can prevent a robot from moving forward.  Vision-based methods therefore learn terrain properties from visual or semantic features \cite{howard2006towards,hadsell2009learning}, rather than treating all geometrically salient structures as obstacles.

The relevance of a visual feature nevertheless depends on the robot and its interaction with the terrain.  Self-supervised traversability methods use robot motion and proprioceptive measurements to generate supervision signals, allowing visual predictions to reflect platform-specific traction, mobility, or collision outcomes~\cite{wellhausen2019should,kahnBADGRAutonomousSelfSupervised2020,gasparino2022wayfast,mattamala2025wild}.  Anomaly detection, proprioceptive feedback, and recovery policies further address situations in which a visual estimate is unreliable or an obstacle has been missed \cite{wellhausen2020safe,ji2022proactive,fuCouplingVisionProprioception2022,jinResilientLeggedLocal2023,xu2025bw}.  For hazards that can lead to irrecoverable states, however, the interaction used to generate a label may high-center or trap the robot and require manual recovery.  Hazard events are therefore time-consuming to collect and remain scarce.

\subsection{Learning from Sparse Failure Data}
With only a small number of hazard events, a classifier can fit the observed scenarios without learning the physical structures associated with failure.  A binary hazard label indicates association with a collision, entrapment, or high-centering event, but it does not identify which image region contains the relevant structure.  Direct training from scratch or fine-tuning with only these labels can therefore rely on texture, illumination, or background correlations that do not transfer to a new environment~\cite{geirhos2020shortcut}. Reweighting objectives such as focal loss increase the contribution of rare positive examples~\cite{lin2017focal}, but do not identify which image regions contain interaction-relevant structure.

Few-shot and meta-learning methods improve adaptation when related tasks are available during meta-training~\cite{qiu2024learning,xue2025fsl,wang2026masked}.  Hazard identification does not satisfy this assumption. Whether a log, root, branch, or vegetation is hazardous depends on its pose, the terrain, and the platform-specific capability, rather than on a fixed semantic category.  A meta-training collection spanning these cases would require broad coverage of the same hazardous interactions that are costly to collect.  


\subsection{Geometry-Guided Vision Foundation Model Adaptation}

Pretrained vision foundation models are a promising starting point because they provide visual representations without requiring a task-specific model to be trained from scratch.  Self-supervised ViTs provide transferable visual features \cite{caron2021emerging,oquab2024dinov2}, and pretrained DINO-ViT features have been effective for wild visual navigation~\cite{mattamala2025wild}.  Combined with parameter-efficient adaptation methods such as LoRA, visual prompt tuning, and AdaptFormer~\cite{hulora,jia2022visual,chen2022adaptformer}, these models reduce the need to train a task-specific network from scratch and make adaptation more feasible when labeled data are limited.  However, their pretraining is generic rather than navigation-specific.  When adapted only with scarce failure labels, a VFM has no explicit preference for visual evidence that is physically relevant to robot--terrain interaction.

Geometry provides a natural way to introduce such a preference for ground robots.  Depth information is widely used in off-road navigation because terrain shape, clearance, and above-ground structure are directly related to navigation \cite{gasparino2022wayfast,wellhausen2020safe}. 
Existing work on geometry-guided VFM adaptation uses geometric information for different purposes. FiT3D uses multi-view geometric consistency to improve the 3D awareness of foundation model features \cite{yue2024improving}, while ViTA combines semantic traversability supervision with geometric knowledge to adapt a VFM for traversability estimation \cite{hwang2026general}.
For hazard identification from rare failures, geometric cues can complement PEFT by biasing adaptation toward physical structures that may affect robot motion rather than incidental appearance cues such as background texture, lighting, or scene layout. This motivates geometry-guided adaptation of vision foundation models for hazard identification from rare failures.

\section{Methodology}

\subsection{Problem Formulation and Overview}
\label{sec:overview}

GAFT uses two complementary data sources.  The first is a traversal dataset without hazard annotations, $\mathcal{D}_{u}=\{(x_i,d_i)\}_{i=1}^{N_u}$, containing temporally matched RGB frames $x_i$ and depth maps $d_i$.  The second is a much smaller hazard set $\mathcal{D}_{h}=\{(x_j,y_j)\}_{j=1}^{N_h}$, where $y_j=1$ denotes a hazard-associated frame, $y_j=0$ a normal traversal frame. 

Stage~1 uses $\mathcal{D}_u$ without hazard labels to construct a proximity-weighted geometric prior from depth relative to an estimated ground plane. This prior guides LoRA adaptation of DINOv2-ViT-S/14, while all original backbone parameters are frozen.  Stage~2 freezes the adapted backbone and trains a linear hazard probe on rollout-weighted patch features from $\mathcal{D}_h$. 

\begin{figure}[t]
\centering
\includegraphics[width=\linewidth]{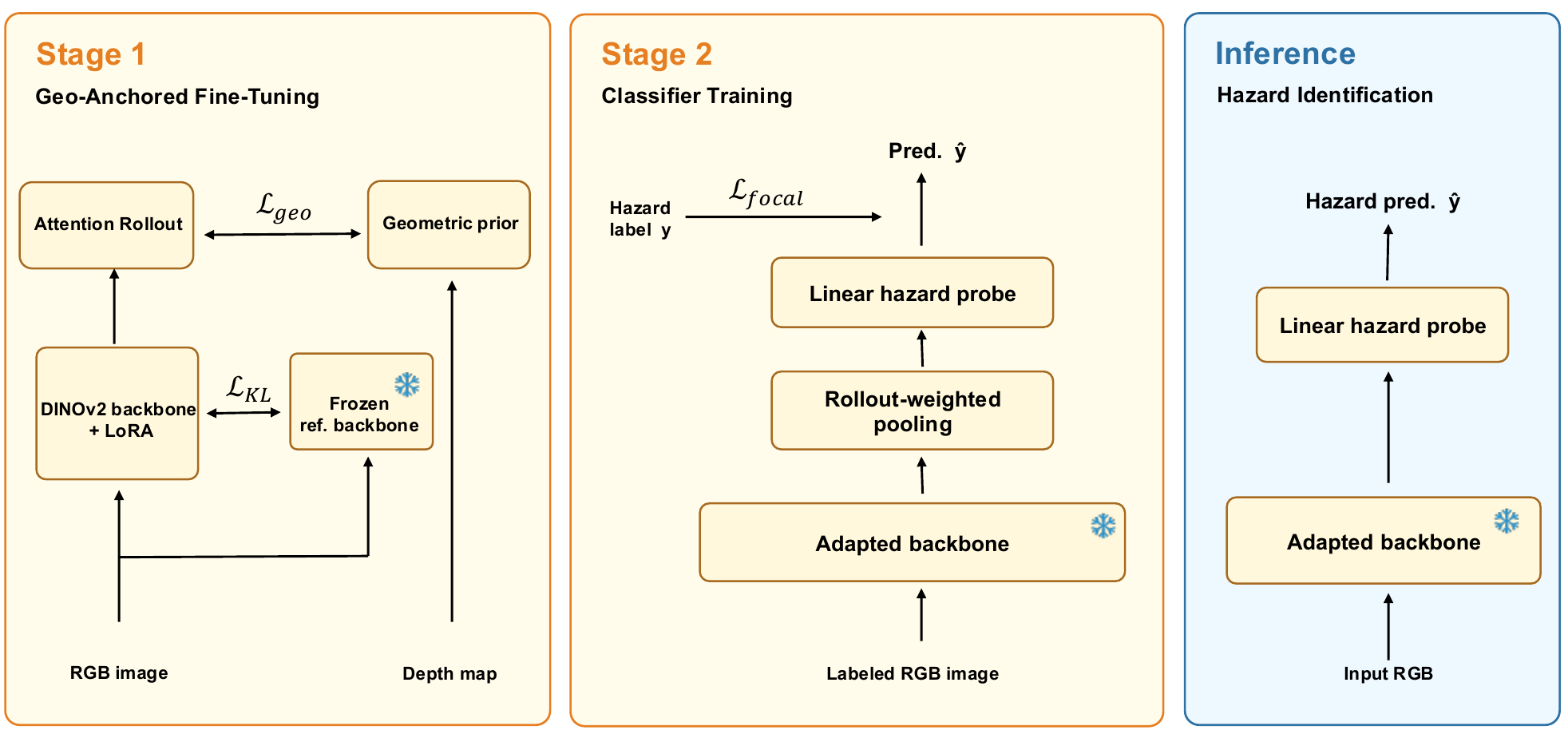}
\caption{GAFT overview.  Stage~1 uses an RGB--depth traversal set without hazard annotations to align a spatial attention-rollout map with a proximity-weighted geometric prior while anchoring the adapted representation to frozen DINOv2.  Stage~2 freezes the adapted backbone and fits a linear hazard probe on scarce labeled RGB frames.  Deployment uses RGB only.}
\label{fig:pipeline}
\end{figure}

\subsection{Proximity-Weighted Geometric Prior}
\label{sec:geom-prior}

For each depth map, valid pixels define a binary mask $W^{\text{valid}}$.  We back-project these pixels into 3D and estimate a ground plane with RANSAC.  Let $\hat{e}(u,v)$ be the signed height of pixel $(u,v)$ above this fitted ground plane, let $d(u,v)$ be its metric depth, and let $d_{\min}$ be the minimum valid depth in the frame.  The geometric target is
\begin{equation}
  \begin{aligned}
  M^*(u,v) ={}& W^{\text{valid}}(u,v)\,
  \exp\!\left[-\lambda\bigl(d(u,v)-d_{\min}\bigr)\right] \\
  &{}\cdot \mathbf{1}\{\epsilon < \hat{e}(u,v) < e_{\max}\}\,
  \mathbf{1}\{d(u,v)<d_{\max}\}.
  \end{aligned}
  \label{eq:mstar}
\end{equation}
The target encodes a proximity-weighted geometric prior: valid pixels within a selected height band above the ground plane receive greater weight when they are closer to the robot, whereas pixels outside this band and distant pixels are suppressed. $M^*$ is bounded in $[0,1]$ by the binary masks and the exponential distance decay. The prior supplies spatial supervision for VFM adaptation, while the hazard classifier is trained on labels derived from robot–terrain interaction outcomes.

\subsection{Geo-Anchored VFM Adaptation}
\label{sec:stage1}

The visual backbone is DINOv2-ViT-S/14.  Its pretrained parameters remain frozen; GAFT adds trainable LoRA adapters to the combined \emph{QKV} projection and, for the QKV+MLP variant, to the feed-forward linear projections in every transformer block.  We evaluate QKV-only and QKV+MLP as two adapter-scope configurations.

GAFT aligns the \texttt{[CLS]}-to-patch attention rollout \cite{abnar2020quantifying} with the geometric prior.  Attention rollout aggregates residual-adjusted attention across transformer layers and provides a differentiable spatial map whose relative patch weights are matched to the geometric prior.  For each block, attention weights $\mathcal{A}_l$ are averaged over heads, mixed with the identity residual connection, row-normalized, and multiplied in layer order:
\begin{equation}
  \begin{aligned}
  \mathbf{a}
      &= \left(\prod_l
      \operatorname{rownorm}\!\left(0.5 \mathcal{A}_l+0.5 I\right)
      \right)_{0,1:},\\
  R &= \operatorname{norm}_{[0,1]}\!\left(
      \operatorname{reshape}(\mathbf{a})\right).
  \end{aligned}
  \label{eq:rollout}
\end{equation}
Here $\mathbf{a}$ contains the raw CLS-to-patch rollout weights, and $R$ is the per-image min--max normalized patch-level map.  During training, $R$ is bilinearly upsampled to the resolution of $M^*$.  Consequently, $\mathcal{L}_{\text{geo}}$ constrains the relative spatial distribution of the rollout rather than its absolute magnitude.

The adaptation objective is
\begin{equation}
  \mathcal{L} = \mathcal{L}_{\text{geo}} + \beta_{kl}\,\mathcal{L}_{\text{KL}}
  \label{eq:loss}
\end{equation}
with a depth-weighted MSE alignment term
\begin{equation}
  \mathcal{L}_{\text{geo}} = \frac{\sum_{i,j} W^{\text{valid}}_{ij}\,
      \bigl(R_{ij} - M^*_{ij}\bigr)^2}
      {\sum_{i,j} W^{\text{valid}}_{ij}}
  \label{eq:geo}
\end{equation}
and a KL representation anchor to the frozen reference backbone:
\begin{equation}
  \mathcal{L}_{\text{KL}} =
      D_{\text{KL}}\!\left(
        \sigma(\mathbf{z}^{\text{frozen}}_{\texttt{CLS}}) \;\big\|\;
        \sigma(\mathbf{z}^{\text{adapt}}_{\texttt{CLS}})
      \right)
  \label{eq:kl}
\end{equation}
where $\mathbf{z}_{\texttt{CLS}}$ denotes the DINOv2 \texttt{[CLS]} embedding and $\sigma$ applies softmax across its feature dimensions.  The KL term is separate from the geometric alignment: the frozen backbone serves as the reference, and the adapted backbone is penalized when its softmax-normalized CLS embedding diverges from the frozen one.  This representation anchor discourages large changes in the CLS embedding while still allowing LoRA to adjust the spatial routing induced by the geometric objective.

\subsection{Hazard Probe and RGB-Only Inference}
\label{sec:stage2}

After adaptation, the DINOv2 backbone and LoRA adapters are frozen.  For each RGB frame, the unnormalised CLS-to-patch rollout weights from Eq.~\ref{eq:rollout} pool the final patch tokens: 
\begin{equation}
  \mathbf{h}(x)=\sum_{p=1}^{P}
      \frac{a_p}{\sum_q a_q}\,\mathbf{t}_p ,
  \label{eq:pool}
\end{equation}
where $P$ is the number of image patches and $\mathbf{t}_p\in\mathbb{R}^{L}$ is the final DINOv2 patch token.  A linear probe maps $\mathbf{h}(x)$ to a hazard logit and is trained on the labeled hazard set with binary focal loss~\cite{lin2017focal}. 

\section{Experimental Evaluation}

\begin{figure}[t]
\centering
\includegraphics[width=0.24\linewidth]{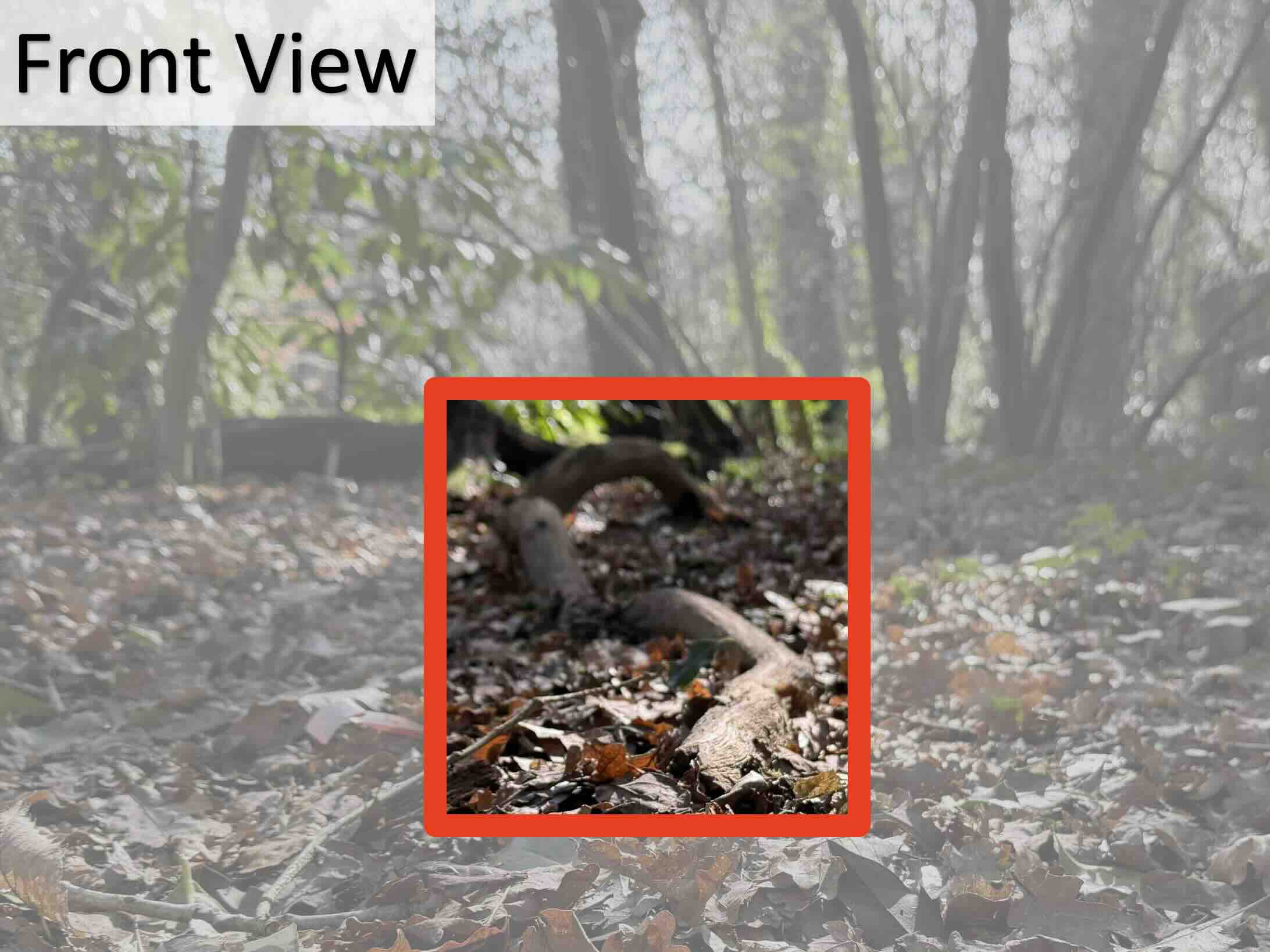}\hfill
\includegraphics[width=0.24\linewidth]{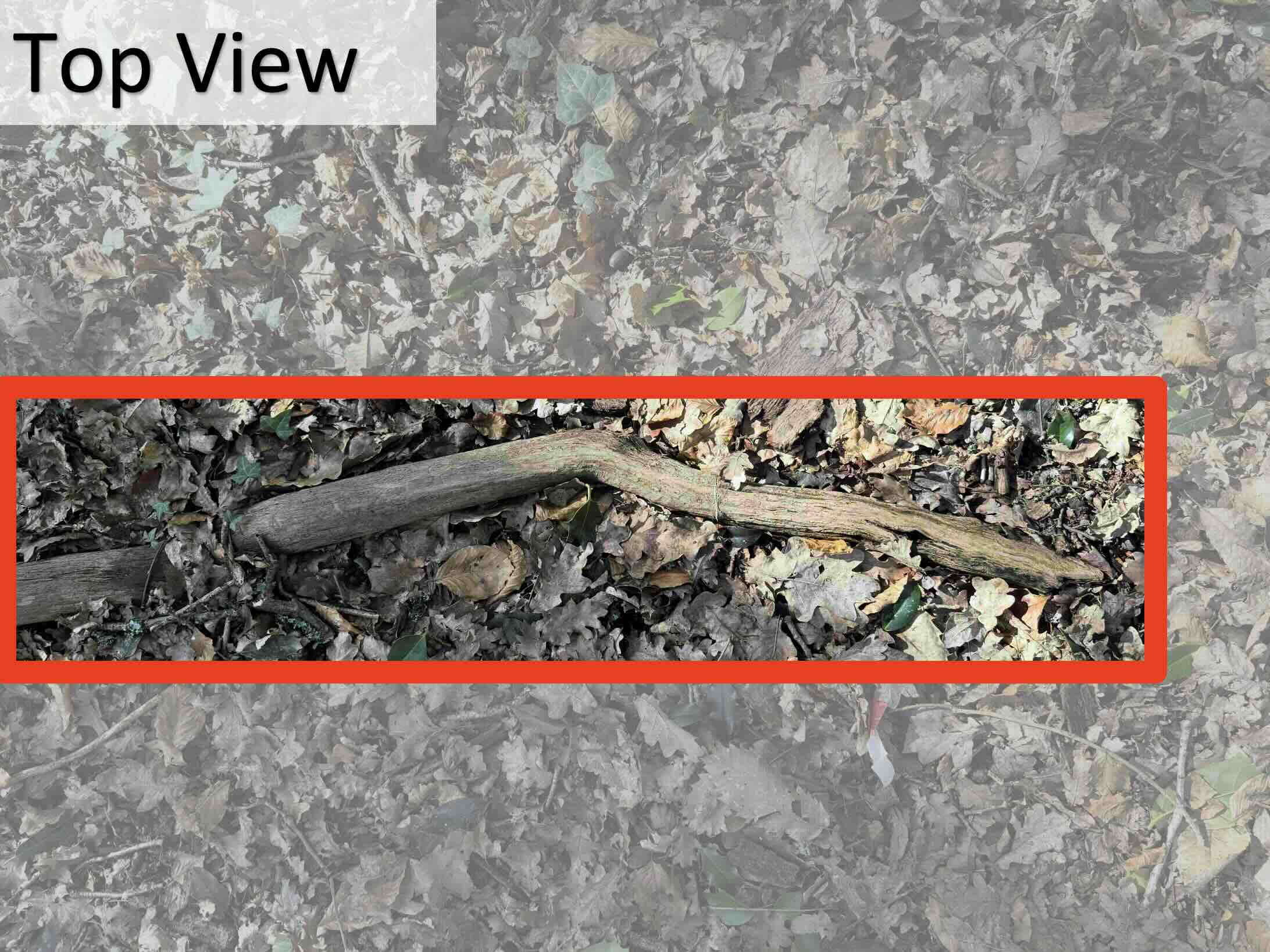}\hfill
\includegraphics[width=0.24\linewidth]{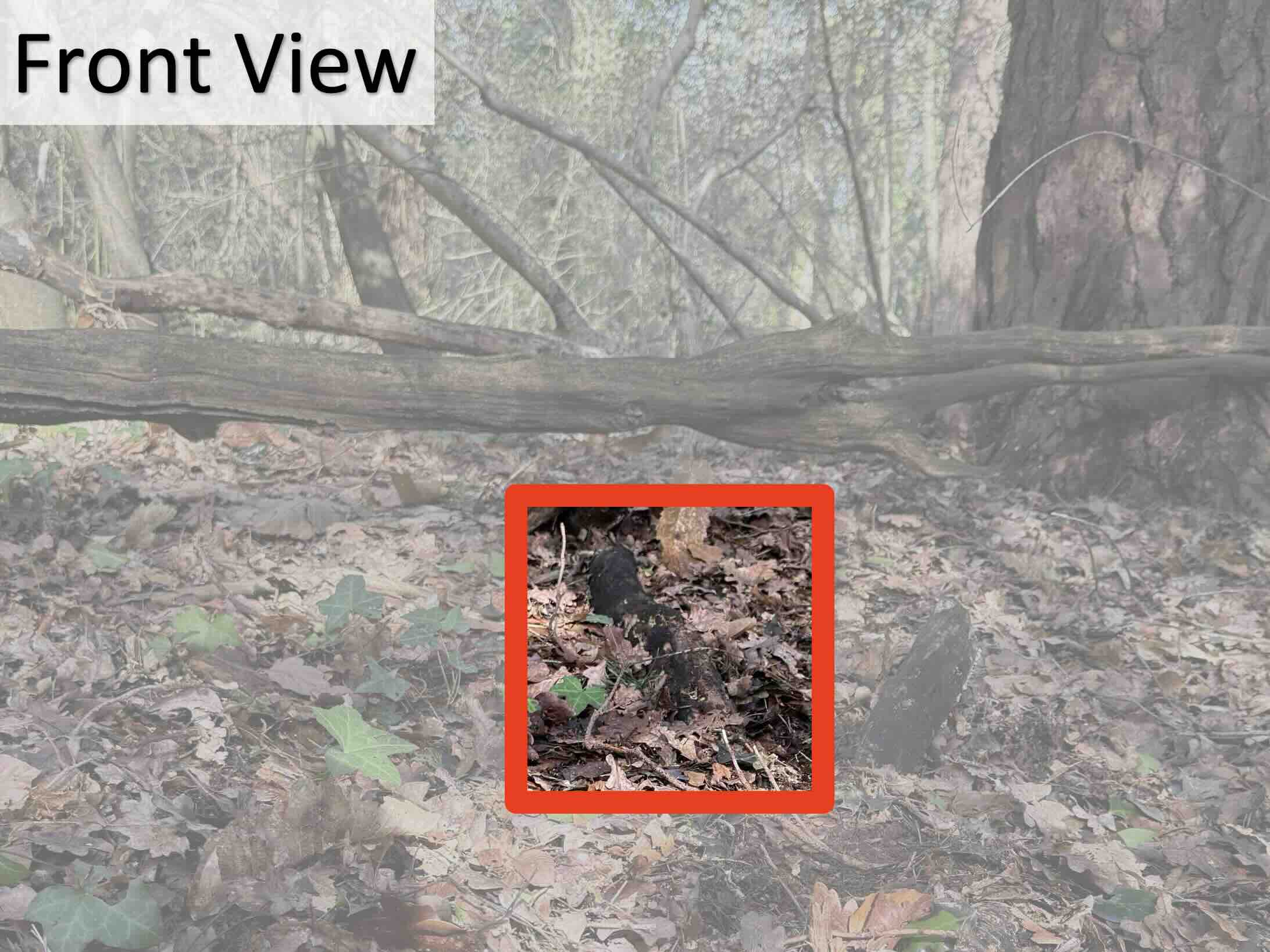}\hfill
\includegraphics[width=0.24\linewidth]{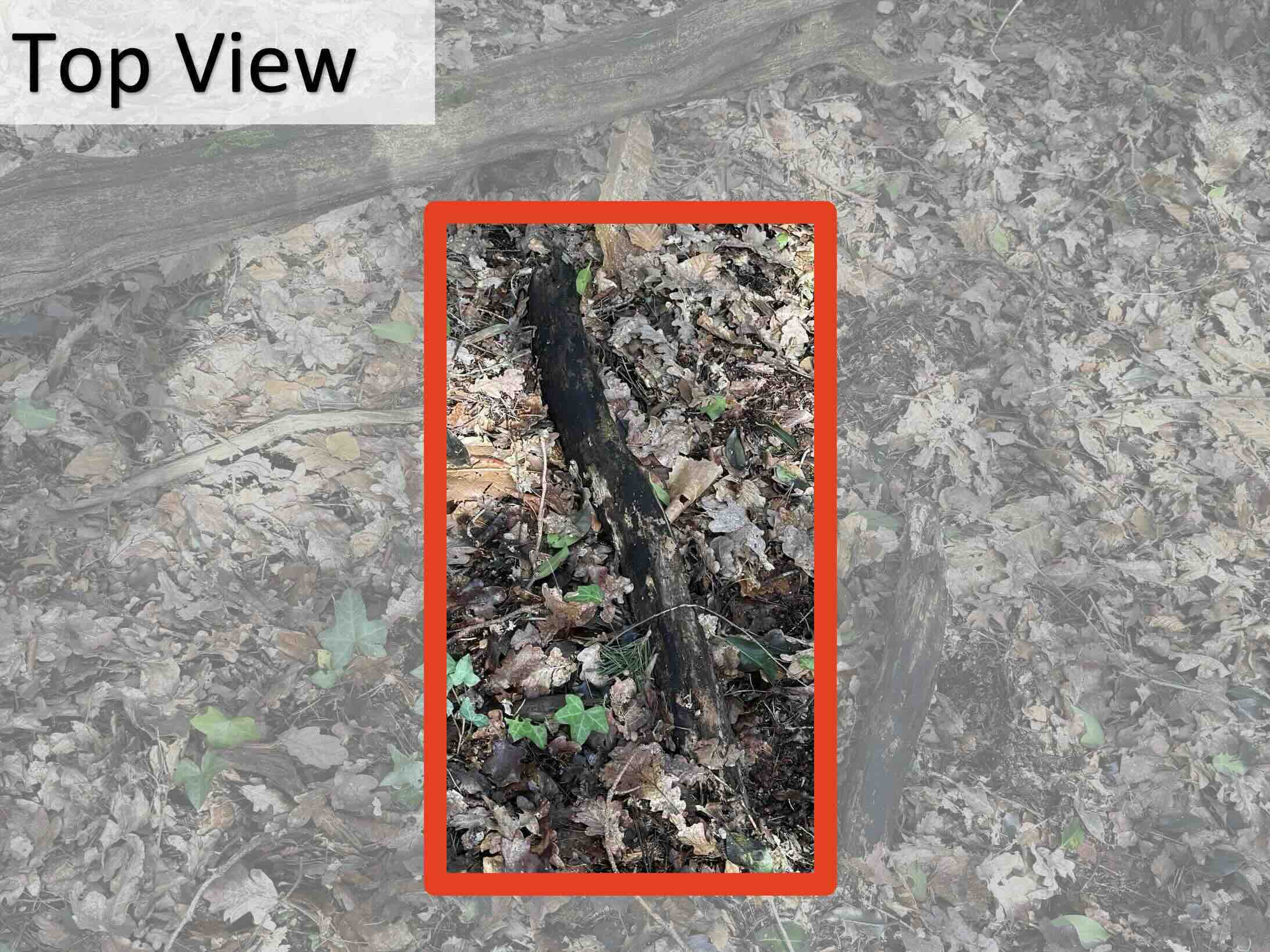}

\vspace{1pt}
\includegraphics[width=0.24\linewidth]{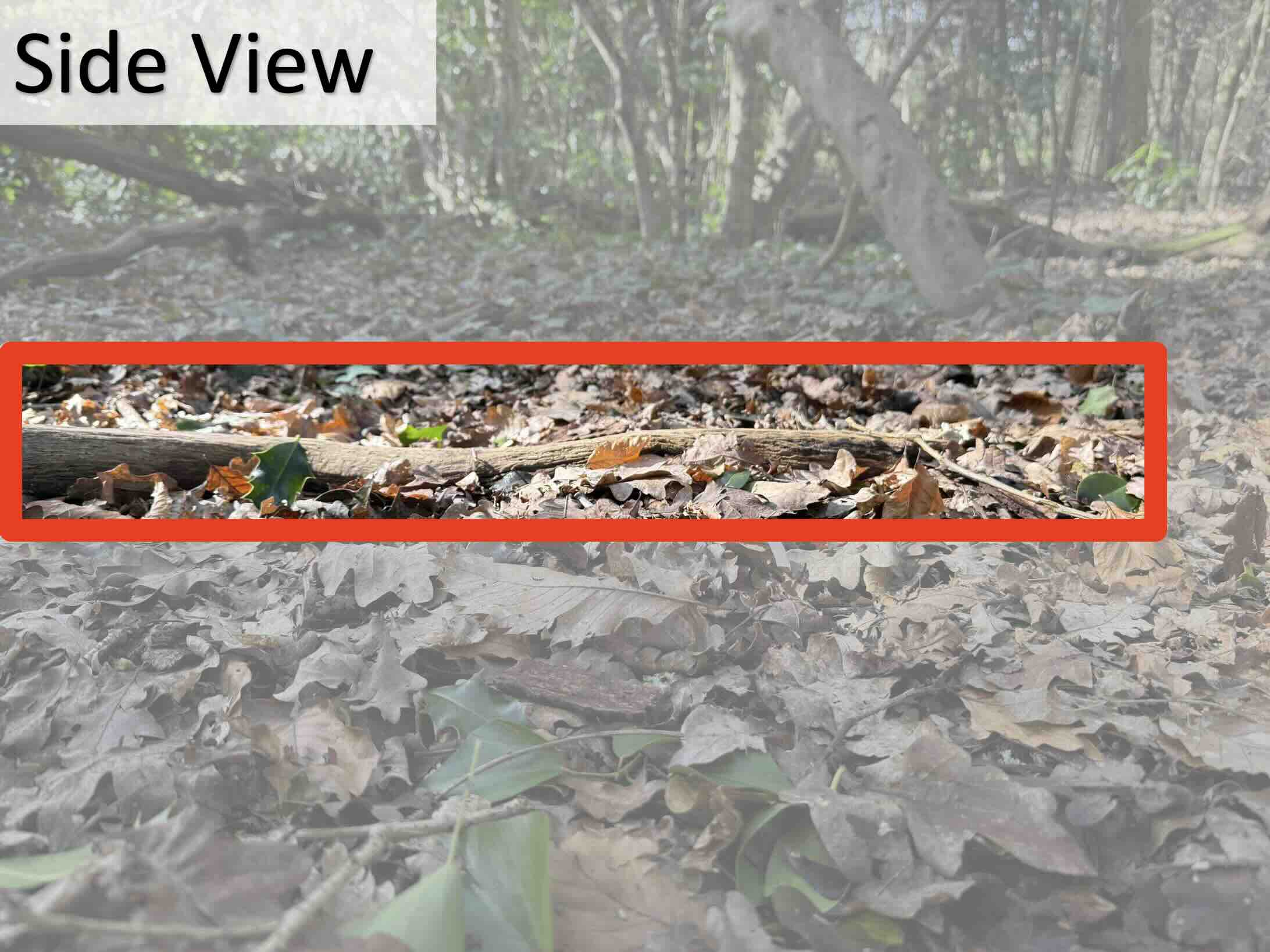}\hfill
\includegraphics[width=0.24\linewidth]{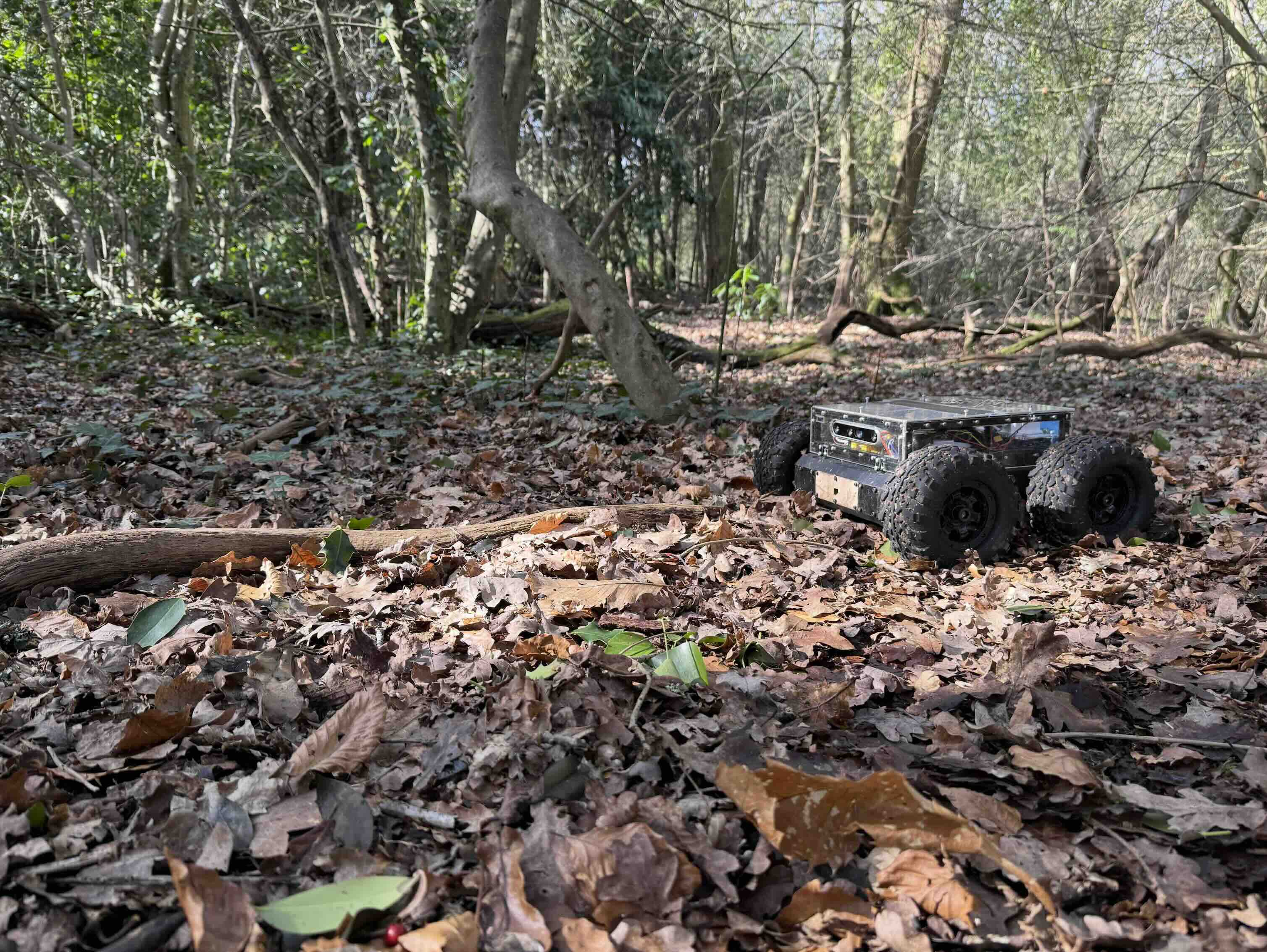}\hfill
\includegraphics[width=0.24\linewidth]{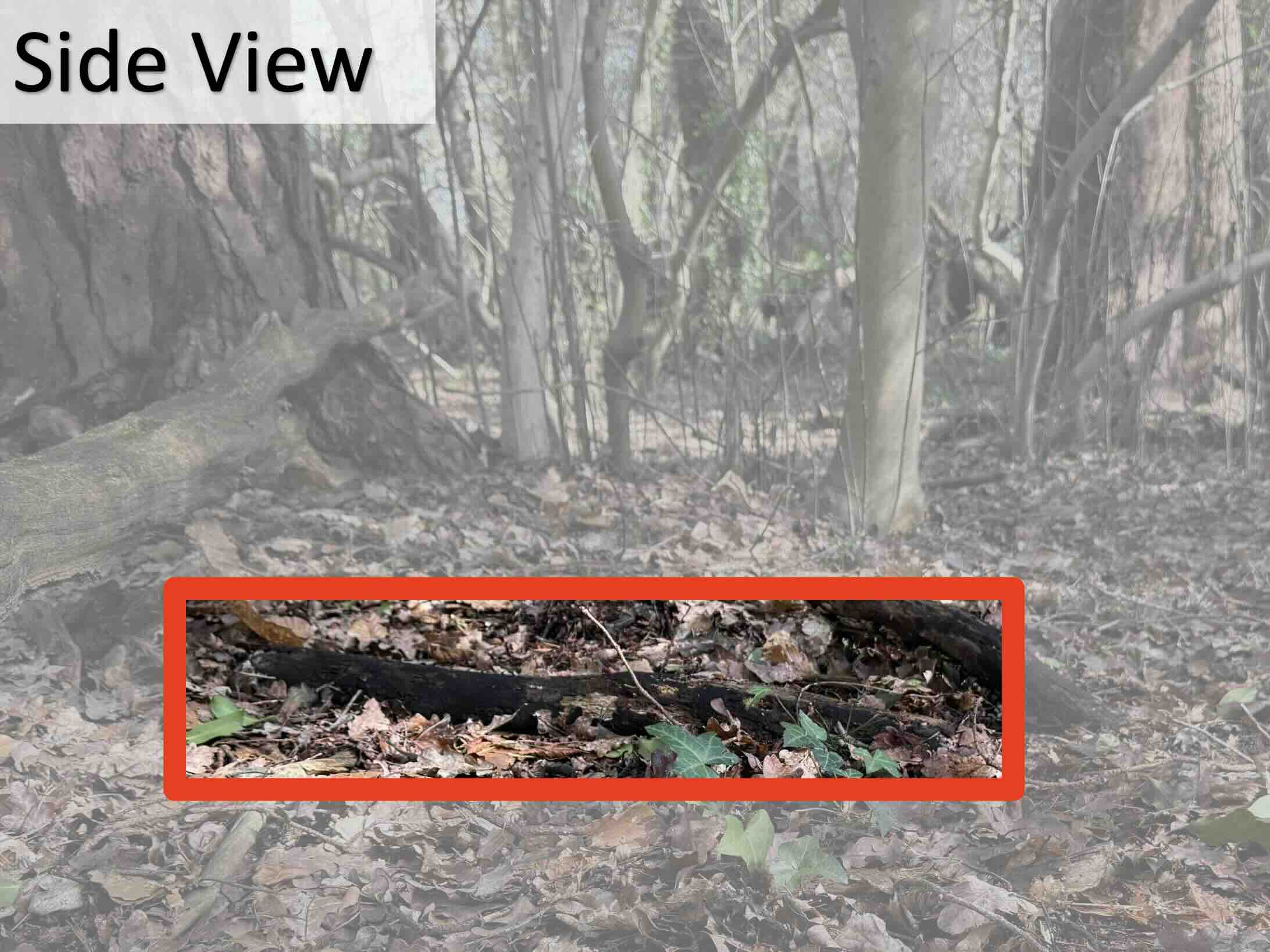}\hfill
\includegraphics[width=0.24\linewidth]{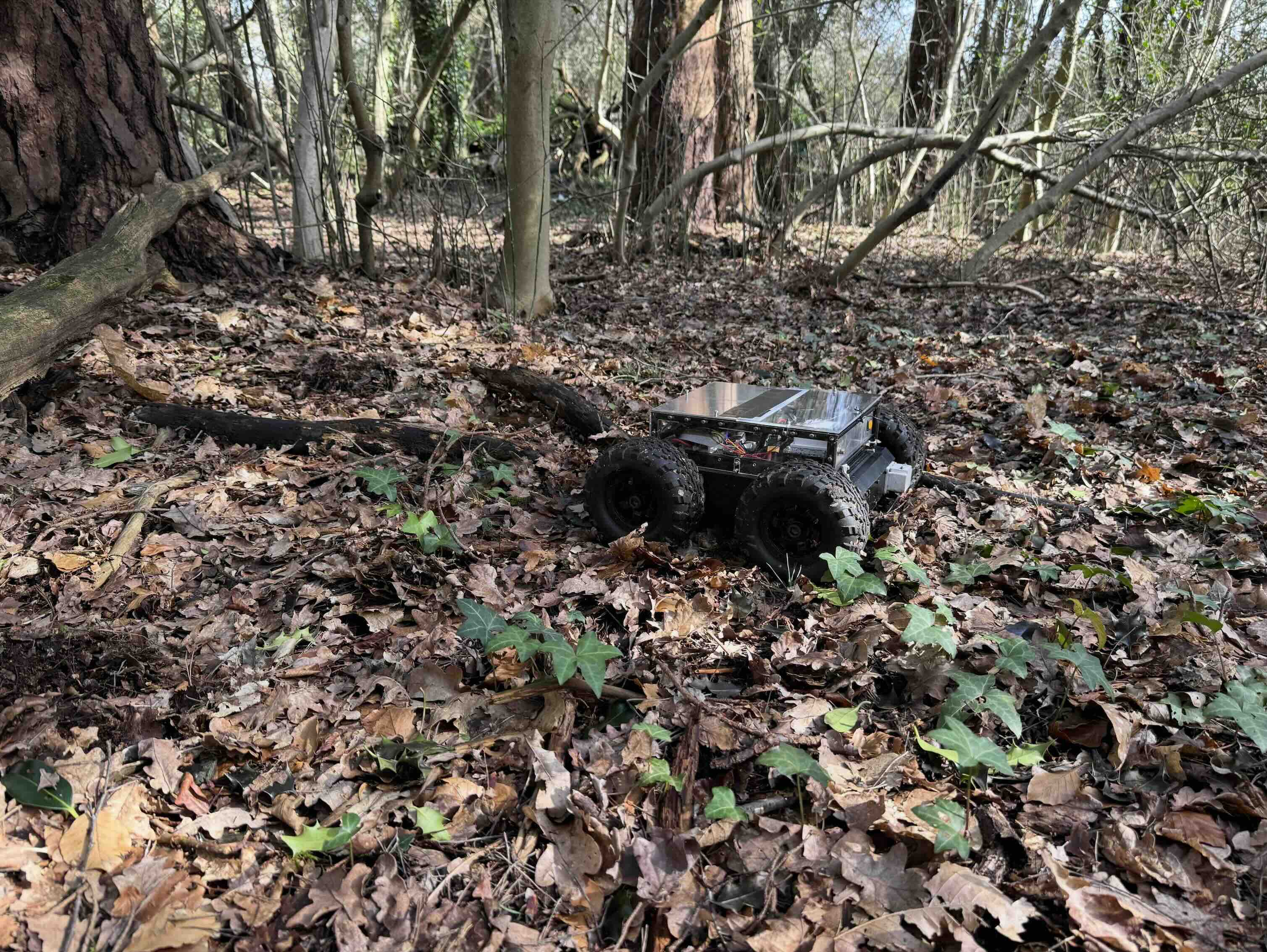}

\caption{Intervention-verified hazards in the RGB benchmark.  The first two columns show an elongated, light-coloured fallen log; the last two show a stubby, dark-coloured woody protrusion.  The top row gives front and top views; the bottom row gives side and rover-scale views.  Red boxes are visual annotations only and are not used for training supervision.}
\label{fig:hazard-scenarios}
\end{figure}

\subsection{Datasets and Protocol}

\paragraph{Data sources.}
The geo-anchored adaptation stage uses 15,223 RGB--depth traversal pairs from ForEnt~\cite{kirdwichai2026forent}, collected by a Go2 legged robot without hazard labels.  Hazard identification uses a separately collected, RGB-only benchmark from a wheeled rover.  The corpora share the broad forest navigation domain but differ in robot platform, camera placement, collection sessions, and failure mechanism.  This mismatch is intentional: GAFT uses depth only to obtain a generic geometric bias toward nearby above-ground structure, not a failure-mode-specific hazard mask. The two corpora are disjoint: no Stage~2 frame, label, scenario, or depth map is used during Stage~1 adaptation.


\paragraph{Event-based hazard collection.}
The rover executed the probing-driven Blind-Wayfarer policy in unstructured forest terrain~\cite{xu2025bw}, while its RGB stream was logged continuously. A hazard event was recorded only when contact with an obstacle immobilized the rover and required human intervention.  We retrospectively extract the associated pre-intervention RGB window as failure-associated evidence; normal frames come from uninterrupted traversal.  Thus the positive label denotes an observed robot--terrain failure event rather than a hand-drawn obstacle category or pixel-level annotation.  
The 124 positive frames were collected from 40 intervention-verified approach runs, each contributing one pre-intervention sequence. Of these, 29 runs involved elongated, light-coloured fallen logs and 11 involved stubby, dark woody protrusions, yielding 67 and 57 positive frames, respectively. The two scenarios differ in colour, shape, and local texture, making LOSO transfer a deliberately hard test of reliance on appearance-specific cues. Figure~\ref{fig:hazard-scenarios} shows their physical scale and visual ambiguity against leaf litter.

\begin{table}[t]
\caption{Data used for geo-anchored adaptation and hazard identification.}
\label{tab:data}
\centering
\small
\begin{tabular}{lcc}
\toprule
\textbf{Corpus} & \textbf{Platform / modality} & \textbf{Supervision / role} \\
\midrule
ForEnt & Go2 / RGB--depth & Adaptation; no hazard labels; 15\,223 pairs \\
Hazard set & Rover / RGB & Recovery labels; 124 hazard, 2\,817 normal \\
\bottomrule
\end{tabular}
\end{table}

\paragraph{Repeated LOSO protocol.}
For each fixed Stage~1 adapter, we run 50 repeated two-direction Leave-One-Scenario-Out (LOSO) evaluations.  Each repetition randomly partitions the original normal frames into two disjoint halves and evaluates both scenario directions.  In either direction, all positives from one scenario, together with one normal half, form the outer training set; original positives from the other scenario and the remaining normal half form the outer test set.  The roles of the scenarios and normal halves are reversed in the complementary direction. Thus, the hazard scenarios and their positive frames remain fixed across repetitions; only the normal-frame partition, inner folds, and probe fitting are redrawn.

Within every outer training set, grouped five-fold CV selects the $F_2$-optimal decision threshold.  Twenty augmented variants are used only for each positive training frame; variants of the same source frame remain in the same inner-CV group, and all test sets contain original frames only.  The affine probe is reinitialized and trained on the complete outer training set before evaluation.  We average the two directional test scores to obtain one repetition score.  Its variation reflects Stage~2 resampling and probe-fitting variability.  For GAFT, we repeat this complete Stage~2 protocol for each independently seeded Stage~1 adapter.  These seeds change the LoRA initialization and subsequent geometry-adaptation trajectory; variation across adapter-run means therefore reflects Stage~1 adaptation, whereas variation within an adapter run reflects Stage~2 evaluation.

\paragraph{Primary metric.}
We report the standard $F_\beta$ score with $\beta\!=\!2$, which weights recall four times more than precision.  This reflects the asymmetric cost of hazard detection: a missed hazard can cause mechanical failure, while a false alarm causes a conservative stop.  At the observed 1:23 positive-to-negative ratio, a stratified-random predictor has expected $F_2\!\approx\!0.042$, which we use as a chance-level reference.

\paragraph{Baselines.}
The \emph{DINOv2-ViT-S (frozen)} baseline uses rollout-weighted features from the frozen backbone with a linear head and represents the unmodified foundation model without task-specific adaptation.  The supervised PEFT baselines adapt DINOv2 directly from scarce hazard labels without Stage~1 geometry: \emph{Sup-LoRA QKV}, \emph{Sup-LoRA QKV+MLP}, deep visual prompt tuning (VPT), and AdaptFormer.  Sup-LoRA QKV+MLP uses the same QKV+MLP adapter scope as the reported GAFT model, testing whether the gain can be explained by adapter capacity alone.  \emph{GAFT QKV} applies geo-anchored adaptation with LoRA on QKV projections only.  \emph{GAFT QKV+MLP} is the reported configuration and also adapts the MLP projections.  All learned methods use the same DINOv2 backbone, the same hazard-probe protocol, and RGB-only inference.

\paragraph{Implementation details.}
All RGB--depth pairs use a bottom-centre $448\times448$ crop resized to $224\times224$, with camera intrinsics adjusted before back-projection.  The RANSAC plane is seeded from the lower $15\%$ of image rows.  Unless otherwise stated, the prior uses $\epsilon=0.05$\,m, $e_{\max}=0.5$\,m, $\lambda=0.3$, and $d_{\max}=5$\,m.  These geometric parameters were fixed before hazard-label training and were not selected on the LOSO test scenarios.  The reported GAFT QKV+MLP model uses LoRA rank $r=32$, $\alpha=0.5$, and $\beta_{kl}=0.05$, for 2.06M trainable adapter parameters.  The Stage~2 probe uses focal loss with $\alpha=0.75$ and $\gamma=2.0$.

\subsection{Main Results}

\begin{figure}[t]
\centering
\includegraphics[width=\linewidth]{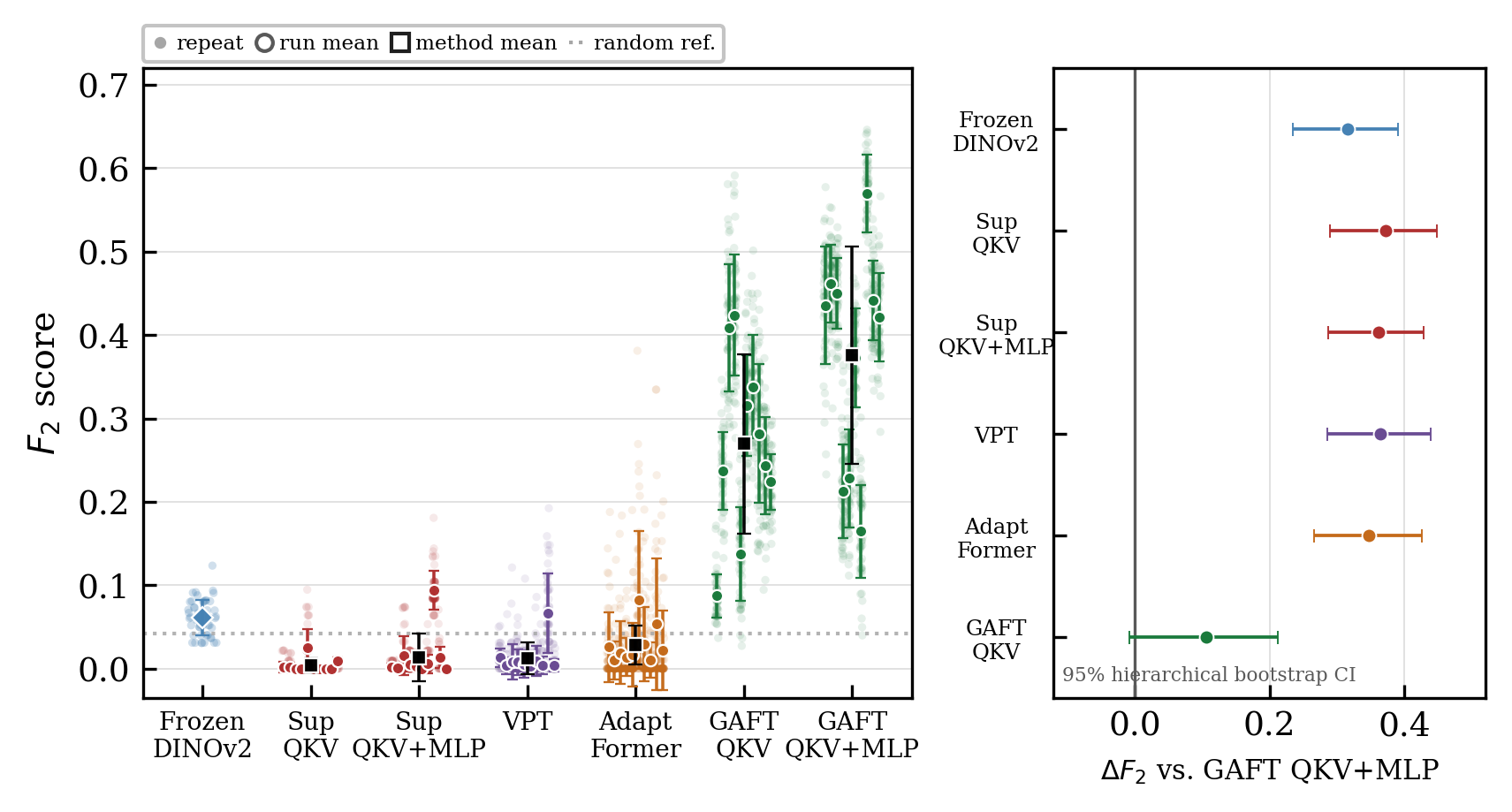}
\caption{Repeated-LOSO results.  Left: a faint point is one Stage~2 repetition, averaged over the two complementary LOSO directions.  For each adapted method, one large coloured circle and bar give the mean $\pm$ standard deviation of 50 repetitions with one fixed Stage~1 adapter; their separation shows variation across ten independently trained adapters.  Each black square and bar summarizes the adapter-run means.  Frozen DINOv2 has no adaptation run, so its diamond and bar summarize its 50 repetitions directly.  Right: paired $\Delta F_2$ of GAFT QKV+MLP with hierarchical-bootstrap 95\% confidence intervals.  ``Sup'' abbreviates Sup-LoRA.}
\label{fig:main-results}
\end{figure}

Table~\ref{tab:main} reports aggregate point estimates under this protocol. GAFT achieves mean $F_2 = 0.3757$ across ten independently initialized Stage~1 adapters, compared with $0.0607$ for frozen DINOv2 and an expected $0.042$ for a stratified-random predictor.  GAFT QKV reaches $0.2695$, and the paired analysis below shows that both GAFT variants reliably outperform frozen DINOv2 and supervised PEFT controls.  The supervised PEFT baselines use the same DINOv2 backbone, focal loss, train-only positive augmentation, inner-CV threshold selection, and repeated LOSO protocol as GAFT, but omit the geo-anchored adaptation stage.  Despite this matched protocol, Sup-LoRA QKV+MLP reaches only $F_2=0.0138$, while VPT and AdaptFormer reach $0.0123$ and $0.0282$.  Table~\ref{tab:peft-transfer} explains this gap: supervised PEFT fits the source scenario but transfers poorly to the held-out hazard scenario, so the low test scores are not simply an optimization or threshold-selection failure.

\begin{table}[t]
\caption{LOSO hazard-identification estimates.  Adapted methods average ten initialization seeds; Figure~\ref{fig:main-results} shows repeated-LOSO variation and paired confidence intervals.}
\label{tab:main}
\centering
\small
\begin{tabular}{lcccc}
\toprule
\textbf{Method} & \textbf{Adapter params} & \textbf{$F_2$ $\uparrow$}
  & \textbf{Recall $\uparrow$} & \textbf{Precision $\uparrow$} \\
\midrule
Stratified random            & -- & $0.042$ & $0.042$ & $0.042$ \\
Frozen DINOv2                & none & $0.0607$ & $0.050$ & $0.650$ \\
Sup-LoRA QKV                 & 0.59M & $0.0039$ & $0.003$ & $0.069$ \\
Sup-LoRA QKV+MLP             & 2.06M & $0.0138$ & $0.012$ & $0.119$ \\
VPT                          & 0.05M & $0.0123$ & $0.012$ & $0.086$ \\
AdaptFormer                  & 0.60M & $0.0282$ & $0.035$ & $0.167$ \\
\midrule
GAFT QKV                      & 0.59M & $0.2695$ & $0.239$ & $0.679$ \\
\textbf{GAFT QKV+MLP}         & 2.06M & $\mathbf{0.3757}$
                              & $\mathbf{0.358}$ & $\mathbf{0.749}$ \\
\bottomrule
\end{tabular}
\end{table}

\begin{table}[t]
\caption{Diagnostic for supervised PEFT: within-scenario fit versus cross-scenario transfer.  Inner-CV and outer-train scores use the source scenario of each LOSO direction; the LOSO test score uses its held-out scenario. All values are means over the repeated LOSO evaluations.}
\label{tab:peft-transfer}
\centering
\small
\begin{tabular}{lccc}
\toprule
& \multicolumn{2}{c}{\textbf{Within-scenario}} & \textbf{Cross-scenario} \\
\cmidrule(lr){2-3}\cmidrule(l){4-4}
\textbf{Method} & \textbf{Inner-CV $F_2$} & \textbf{Outer-train $F_2$} & \textbf{LOSO test $F_2$} \\
\midrule
Sup-LoRA QKV      & 0.944 & 0.975 & 0.0039 \\
Sup-LoRA QKV+MLP  & 0.952 & 0.987 & 0.0138 \\
VPT               & 0.853 & 0.922 & 0.0123 \\
AdaptFormer       & 0.832 & 0.860 & 0.0282 \\
\bottomrule
\end{tabular}
\end{table}

Table~\ref{tab:peft-transfer} shows that the poor PEFT scores in Table~\ref{tab:main} are not simply explained by failure to fit the source scenario.  Every supervised PEFT method obtains high inner-CV and outer-train $F_2$ within that scenario, but low $F_2$ when LOSO evaluates the other scenario.  On the held-out scenario, these models predict hazard for only 0.05--1.51\% of frames and have zero recall in 70.2--90.2\% of evaluation folds.  This pattern is consistent with scenario-specific cue reliance under scarce failure labels, rather than with a simple optimization failure. It does not by itself identify the cues or establish their causal role.

To test whether the gains persist under matched Stage~2 evaluations, we pair methods on the same LOSO direction and repetition.  For each Stage~1 adapter run, paired differences are averaged over its repeated outer evaluations; hierarchical bootstrap confidence intervals then resample adapter runs and paired evaluations.  Across ten seed-matched runs (1,000 paired Stage~2 folds per comparison), Figure~\ref{fig:main-results} shows positive intervals against frozen DINOv2 ($\Delta F_2=+0.3150$ $[0.2350,0.3904]$) and matched-capacity Sup-LoRA QKV+MLP ($+0.3619$ $[0.2865,0.4284]$), but the advantage over GAFT QKV is not robust ($+0.1062$ $[-0.0073,0.2118]$).

Figure~\ref{fig:main-results} separates the two levels of variation.  Each coloured circle and bar gives the mean and standard deviation of 50 Stage~2 repeats for one independently initialized adaptation run; for GAFT, this is one Stage~1 adapter.  The black square and bar summarize the resulting run means. The supervised PEFT baselines remain near zero at both levels, whereas GAFT shifts the distribution upward.  In this setting, matched adapter capacity alone does not yield cross-scenario performance; coupling it with the geo-anchored objective does.  MLP adaptation should nevertheless be interpreted as a capacity choice rather than a statistically isolated contribution.


The broad GAFT range in Figure~\ref{fig:main-results} should not be read as instability of a fixed adapted backbone under repeated LOSO.  Holding each Stage~1 adapter fixed, we perform 50 independent Stage~2 repetitions; the resulting standard deviation, averaged across the ten independently trained adapters, is $0.0575$ for GAFT QKV and $0.0536$ for GAFT QKV+MLP.  The wider separation among the corresponding adapter means is therefore introduced when Stage~1 constructs the geometry-adapted backbone.  In this implementation, a Stage~1 seed affects both LoRA initialization and the subsequent adaptation trajectory, including minibatch order and checkpoint selection, which are known sources of fine-tuning variability \cite{dodge2020fine,hayou2024impact}.

\subsection{Ablation Studies}
\label{sec:ablation}

We first test whether the pooling used by the frozen reference explains its weak performance, then examine how KL anchoring strength and LoRA capacity affect performance under the geo-anchored objective.

\begin{table}[t]
\caption{Pooling control and ablations.  Frozen pooling results are mean $\pm$ standard deviation over 50 Stage~2 repetitions.  GAFT ablations are mean $\pm$ standard deviation across five Stage~1 seeds.}
\label{tab:ablation}
\centering
\small
\begin{tabular}{llc}
\toprule
\textbf{Factor} & \textbf{Setting} & \textbf{$F_2$ $\uparrow$} \\
\midrule
\multirow{3}{*}{Frozen pooling}
  & CLS & $0.0095 \pm 0.0170$ \\
  & Mean patch & $0.0588 \pm 0.0356$ \\
  & Rollout & $0.0607 \pm 0.0209$ \\
\midrule
\multirow{4}{*}{$\beta_{kl}$, QKV/RANSAC}
  & 0.001 & $0.0596 \pm 0.0624$ \\
  & 0.005 & $0.0895 \pm 0.0914$ \\
  & 0.010 & $0.0529 \pm 0.0118$ \\
  & 0.050 & $0.2587 \pm 0.1535$ \\
\midrule
\multirow{3}{*}{Rank, QKV/RANSAC, $\beta_{kl}=0.05$}
  & $r=8$  & $0.0763 \pm 0.0527$ \\
  & $r=16$ & $0.2369 \pm 0.0238$ \\
  & $r=32$ & $0.2587 \pm 0.1535$ \\
\bottomrule
\end{tabular}
\end{table}

\paragraph{Pooling control.}
For frozen DINOv2, CLS pooling performs poorly, whereas mean-patch and rollout pooling yield similar $F_2$ values (Table~\ref{tab:ablation}).  Thus the weak frozen reference is not an artifact of rollout pooling.  We retain rollout pooling because it provides the spatial interface through which GAFT applies its Stage~1 geometric supervision.

\paragraph{KL anchoring.}
In the QKV-only RANSAC setting, the strong-anchor configuration $\beta_{kl}=0.05$ reaches a five-seed mean of $0.2587 \pm 0.1535$, compared with $0.0596 \pm 0.0624$ at $\beta_{kl}=0.001$ (Table~\ref{tab:ablation}).  The weak-anchor settings are highly seed-sensitive and do not consistently exceed the frozen baseline.  The ablation therefore shows that performance depends on sufficiently strong anchoring, without isolating the particular mechanism by which the KL term provides it.

\paragraph{LoRA capacity.}
We also vary QKV-only LoRA rank at $\beta_{kl}=0.05$ and RANSAC $M^*$.  In this five-seed ablation, $r=8$ is weak and seed-sensitive ($0.0763 \pm 0.0527$), whereas $r=16$ has a higher and more stable mean ($0.2369 \pm 0.0238$).  Rank $r=32$ reaches a slightly higher mean ($0.2587 \pm 0.1535$) but with larger seed variance.  These exploratory results indicate that very low rank is inadequate for this objective, while increasing capacity alone does not guarantee a stable improvement.

\section{Discussion}

The central comparison shows that, in this data-scarce setting, matched adapter capacity and hazard-label supervision alone do not yield cross-scenario hazard identification.  Supervised PEFT fits the source scenario but transfers poorly to the held-out scenario, whereas geo-anchored adaptation substantially improves the repeated-LOSO distribution.  This contrast supports the role of geo-anchored adaptation as an inductive bias: the supervised PEFT controls learn a source-scenario decision rule, whereas GAFT improves transfer to the held-out hazard scenario.

The ablations further show that the benefit is conditional.  Weak KL anchoring and very low LoRA rank yield weak or seed-sensitive results, while the difference between GAFT QKV and GAFT QKV+MLP is not robust under the paired analysis. Accordingly, QKV+MLP is the reported practical configuration rather than a separate statistically isolated contribution.  Alternative representation anchors and more accurate geometric targets remain open design choices.

These results should be interpreted as evidence for offline RGB hazard identification under rare, intervention-verified field failures.  The benchmark contains two verified forest hazard morphologies, so the evaluation measures transfer between these morphologies rather than robustness across arbitrary seasons, platforms, sensor placements, or closed-loop navigation policies.  The 50 repeated LOSO evaluations quantify Stage~2 resampling and probe-fitting variation, not independent environments.  GAFT is also not claimed to produce a pixel-accurate hazard explanation: attention rollout is used as a differentiable spatial-routing interface, and $M^*$ is a proximity-weighted geometric proxy rather than a hazard mask.  Closed-loop validation of whether GAFT reduces immobilization, recovery frequency, or mission time remains future work.

\section{Conclusion}

We introduced GAFT, a two-stage method that uses a proximity-weighted geometric prior to adapt an RGB vision foundation model before hazard-label supervision. The prior is available only during Stage~1; Stage~2 and deployment use RGB alone. Across ten independently initialized adaptations and repeated LOSO evaluation, GAFT improves frozen DINOv2 from mean $F_2=0.0607$ to $0.3757$.  Matched supervised PEFT controls fit their source scenario but fail to transfer reliably to the held-out one.  Together, these results support a narrow conclusion: geo-anchored adaptation can improve RGB-only hazard identification when hazard labels are scarce and adaptation remains anchored to pretrained features.


\bibliographystyle{splncs04}
\bibliography{main}

\end{document}